%% file: main.tex
\pdfoutput=1
\documentclass[10pt,twocolumn,letterpaper]{article}
\usepackage[pagenumbers]{cvpr}
\input{preamble}

\definecolor{cvprblue}{rgb}{0.21,0.49,0.74}
\usepackage{xcolor}
\usepackage{listings}
\usepackage[pagebackref,breaklinks,colorlinks,allcolors=cvprblue]{hyperref}
\hypersetup{pdfauthor={Chenzhi Liu, Jiehong Lin, Wei Huang, Xuzhao Huang, Haoyuan Li, Yang Zhang, Junchao Huang, Chenyeyang Cai, Bo Wang, Kazuhiro Kosuge, Zhongrui Wang, Xiaojuan Qi},pdftitle={RoboEvo2: Recursive Robot Self-Improvement via Harness-Policy Co-Evolution}}
\def\confName{CVPR}
\def\confYear{2027}
\makeatletter
\patchcmd{\@maketitle}{\vskip .375in}{\vskip .1in}{}{\PackageWarning{arxiv}{title patch failed}}
\patchcmd{\@maketitle}{\vspace*{12pt}}{\vspace*{-20pt}}{}{\PackageWarning{arxiv}{title patch failed}}
\patchcmd{\@maketitle}{\vspace*{24pt}}{\vspace*{12pt}}{}{\PackageWarning{arxiv}{title patch failed}}
\makeatother

\title{RoboEvo$^2$: Recursive Robot Self-Improvement via Harness--Policy Co-Evolution}
\author{%
Chenzhi Liu$^{1,*}$\quad
Jiehong Lin$^{1,*,\dagger}$\quad
Wei Huang$^{1}$\quad
Xuzhao Huang$^{1}$\\
Haoyuan Li$^{2}$\quad
Yang Zhang$^{3}$\quad
Junchao Huang$^{4}$\quad
Chenyeyang Cai$^{2}$\\
Bo Wang$^{1}$\quad
Kazuhiro Kosuge$^{5}$\quad
Zhongrui Wang$^{3}$\quad
Xiaojuan Qi$^{1,\ddagger}$\\[1pt]
{\normalsize $^{1}$The University of Hong Kong\quad
$^{2}$Zhejiang University\quad
$^{3}$Southern University of Science and Technology}\\
{\normalsize $^{4}$The Chinese University of Hong Kong, Shenzhen\quad
$^{5}$City University of Hong Kong}\\[1pt]
{\small $^{*}$Equal contribution\quad $^{\dagger}$Project leader\quad $^{\ddagger}$Corresponding author: \href{mailto:xjqi@eee.hku.hk}{\texttt{xjqi@eee.hku.hk}}}\\
{\small Project page: \href{https://kaiknower.github.io/RoboEvo2/}{\texttt{kaiknower.github.io/RoboEvo2}}\qquad
Code: \href{https://github.com/kaiknower/RoboEvo2}{\texttt{github.com/kaiknower/RoboEvo2}}}
}

\begin{document}
\setlength{\textfloatsep}{10pt plus 2pt minus 2pt}
\setlength{\dbltextfloatsep}{10pt plus 2pt minus 2pt}
\setlength{\floatsep}{8pt plus 2pt minus 2pt}
\setlength{\dblfloatsep}{8pt plus 2pt minus 2pt}
\maketitle
\begin{strip}
\centering
\captionsetup{type=figure}
\includegraphics[width=0.99\textwidth]{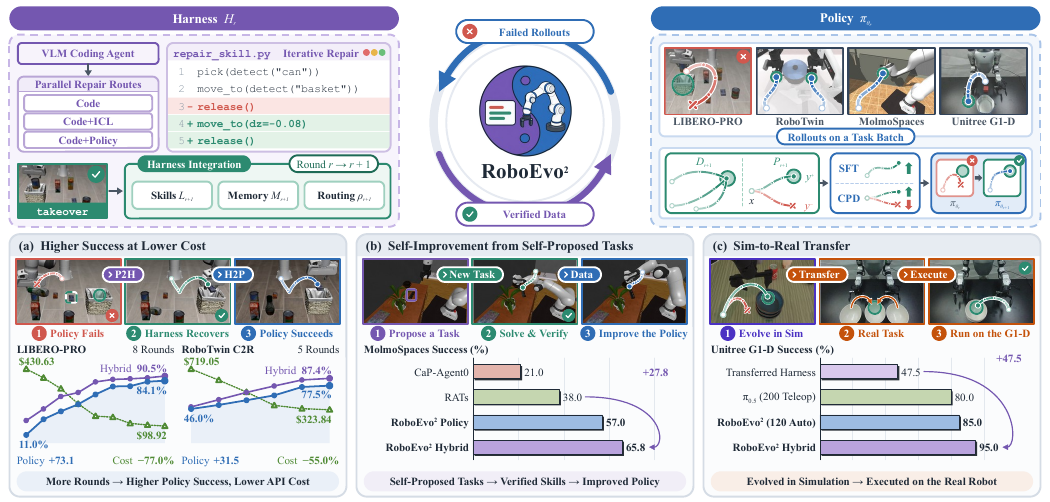}
\caption{
\textbf{Overview of RoboEvo$^2$.}
\textit{Co-evolution loop (top):} the policy is rolled out on a task batch, and failed tasks are
passed to a coding agent, which repairs the harness through parallel Code, Code+ICL, and
Code+Policy routes. Verified repairs are kept as skills, memory, and routing rules, and the
verified trajectories update the policy.
\textit{Results (bottom):} on LIBERO-PRO and RoboTwin C2R, success of the policy alone and with
the harness rises while API cost falls over rounds; in MolmoSpaces, the system improves on
tasks it proposes itself; on the Unitree G1-D, the harness evolved in RoboTwin transfers
to the real robot and collects the demonstrations that train the policy.
}
\label{fig:framework}
\end{strip}
\begin{abstract}
\input{sections/0_abstract}
\end{abstract}
\addtocontents{toc}{\protect\setcounter{tocdepth}{-1}}
\input{sections/1_introduction}
\input{sections/2_related_work}
\input{sections/3_method}
\input{sections/4_experiments}
\input{sections/5_conclusion}
\section*{Acknowledgments}
The work has been supported by Hong Kong Research Grant Council -- General Research Fund Scheme
(Grant No.~17202422, 17212923, 17215025) Theme-based Research (Grant No.~T45-701/22-R), and Strategic
Topics Grant (Grant No.~STG3/E-605/25-N). Part of the described research work is conducted in the JC STEM
Lab of Robotics for Soft Materials funded by The Hong Kong Jockey Club Charities Trust.
\input{sections/6_llm_usage_statement}
\par
{\small
\bibliographystyle{ieeenat_fullname}
\bibliography{references}
}

\clearpage
\setlength{\textfloatsep}{20pt plus 2pt minus 4pt}
\setlength{\dbltextfloatsep}{20pt plus 2pt minus 4pt}
\setlength{\floatsep}{12pt plus 2pt minus 2pt}
\setlength{\dblfloatsep}{12pt plus 2pt minus 2pt}
\appendix
\crefalias{section}{appendix}
\crefalias{subsection}{subappendix}
\crefname{appendix}{Appendix}{Appendices}
\Crefname{appendix}{Appendix}{Appendices}
\crefname{subappendix}{Appendix}{Appendices}
\Crefname{subappendix}{Appendix}{Appendices}
\setlength{\parskip}{0pt plus 4pt}
\addtocontents{toc}{\protect\setcounter{tocdepth}{2}}
\onecolumn
\tableofcontents
\twocolumn
\input{sections/A_supplement}
\end{document}

%% file: preamble.tex
\usepackage[T1]{fontenc}
\usepackage{tabularx,multirow}
\usepackage{cuted}

%% file: sections/0_abstract.tex
Robots should adapt to unfamiliar environments and acquire new skills through interaction, then execute those skills efficiently with practice.
Vision-language-action policies provide efficient control, but often need additional human demonstrations to handle unfamiliar settings; reasoning agents can discover solutions using broad prior knowledge and robot-control APIs, but repeated deliberation and repair are costly.
Inspired by human learning through reasoning and practice, we propose \textbf{RoboEvo$^2$}, a framework for recursive \emph{harness--policy co-evolution}.
In \textbf{Policy-to-Harness (P2H)} evolution, policy failures guide a coding agent to search complementary Code, Code+ICL, and Code+Policy routes, consolidating verified solutions into reusable skills, routing strategies, and experience memory.
In \textbf{Harness-to-Policy (H2P)} evolution, verified trajectories provide supervised training, while \emph{Counterfactual Preference Distillation (CPD)} teaches the policy to favor successful corrections over failures from the same starting state.
The improved policy provides stronger motor skills for subsequent search, while its remaining failures redirect harness evolution, closing the reciprocal loop.
Through recursive co-evolution without additional human demonstrations, RoboEvo$^2$ improves standalone success from 11.0\% to 84.1\% and hybrid success from 30.4\% to 90.5\% on LIBERO-PRO, with similar gains and reduced agentic costs on RoboTwin C2R. Further experiments demonstrate autonomous task acquisition, cross-embodiment harness transfer, and 95.0\% hybrid success on the Unitree G1-D.
Our models and code will be publicly released.

%% file: sections/1_introduction.tex
\section{Introduction}

Robots should adapt rapidly to unfamiliar environments, acquire new skills through interaction, and execute those skills efficiently with practice. Human learning motivates this progression from deliberate problem solving to fluent execution, while retaining the flexibility to tackle new situations. Current robot systems offer complementary strengths toward this goal. Vision-language-action models (VLAs)~\cite{kim2024openvla,black2025pi0,intelligence2025pi05,zheng2025xvla} and world action models (WAMs)~\cite{yuan2026fastwam,guo2026xwam,huang2026longwam} enable efficient visuomotor control, but remain vulnerable to task and scene shifts~\cite{zhou2025liberopro,ross2011dagger}. Adapting them through imitation learning often requires collecting many additional human demonstrations, making continual skill acquisition costly. Foundation-model agents can draw on broad prior knowledge, reason about unfamiliar situations, and invoke robot-control APIs to discover and refine solutions~\cite{liang2023code,fu2026capx,zhang2026harness,zhang2026rats}. However, repeated reasoning, tool use, and repair introduce latency and API costs that limit reliance on agentic execution for routine behavior.

Realizing this progression requires a reciprocal learning loop between an agent's \emph{harness}---its executable skills, memory, and orchestration---and its low-level policy. Harness-centered recursive self-improvement methods refine programs and accumulate experience through feedback~\cite{ding2026zetta,li2026embodiedevo,song2026embodiedrsi,wang2026rpg}, but keeping the motor policy fixed does not transfer newly acquired skills into the learned controller. Policy-centered methods use foundation models to guide autonomous data collection and improve learned controllers~\cite{ha2023scaling,zhou2024soar,liu2026mobiagent}, but do not jointly adapt the harness to the controller's changing abilities. Co-evolution must therefore connect \emph{learning efficient policy behaviors from harness discoveries} with \emph{adapting the harness to the improving policy}. This reciprocity is essential because policy improvement changes both the motor skills available to the agent and the failures that still require exploration and repair.

We introduce \textbf{RoboEvo$^2$}, a framework for recursive \emph{harness--policy co-evolution} that uses current-policy failures to drive both updates (\cref{fig:framework}). In \textbf{Policy-to-Harness (P2H)} evolution, a coding agent diagnoses failures, discovers verified corrections, and consolidates them into reusable skills, routing strategies, and persistent memory. In \textbf{Harness-to-Policy (H2P)} evolution, successful trajectories and matched failure--correction pairs teach the policy to reproduce effective corrections and disfavor failed alternatives. The updated policy then supplies stronger motor skills for subsequent agentic search, while its remaining failures redirect harness refinement. Concurrent studies also investigate harness--policy coupling~\cite{coharness2026,harnessforge2026,wei2026scaleselection,wang2026adarsi}. RoboEvo$^2$ connects failure-guided, multi-route solution discovery with explicit learning from matched failures and corrections, progressively reducing repeated agentic intervention as acquired skills are internalized into the policy.

\textit{First, to discover complementary capabilities, we develop a failure-conditioned, multi-route P2H repair engine.} For each policy failure, it searches three solution routes in parallel: \emph{Code} refines control programs, \emph{Code+ICL} incorporates demonstration guidance, and \emph{Code+Policy} composes programs with learned motor skills. Execution feedback drives iterative repair, and the successful candidate with the lowest normalized motion jerk is selected under matched conditions. This design combines complementary control capabilities to generate verified, smooth corrections targeted at current policy weaknesses.

\textit{Second, to internalize corrective capabilities, we introduce Counterfactual Preference Distillation (CPD).} CPD pairs failed and corrected continuations from the same starting state and optimizes a margin between their action-prediction losses relative to a frozen reference policy. Asymmetric proximity penalties discourage degradation of corrective behaviors and excessive suppression of failed alternatives. Combined with supervised learning on verified trajectories, CPD teaches both which corrections to reproduce and which failed alternatives to disfavor. Failure-aware sampling and continual-learning regularization support targeted acquisition and capability retention.

\textit{Third, to sustain iterative capability reuse, we develop a harness update mechanism coupled with policy reuse across rounds.} Compatible repairs update executable skills and routing logic, while persistent memory retains failure contexts, attempted solutions, and validated guidance. The updated policy then drives subsequent rollouts and participates in Code+Policy search, changing both the failures requiring exploration and the motor primitives available for repair. This closes the reciprocal loop: accumulated harness knowledge supports correction discovery, while internalized policy capabilities directly support subsequent search.

Experiments demonstrate recursive adaptation to distribution shifts without additional human demonstrations. Across eight rounds on LIBERO-PRO~\cite{zhou2025liberopro} and five rounds on RoboTwin C2R~\cite{chen2025robotwin2}, standalone success improves from 11.0\% to 84.1\% and 46.0\% to 77.5\%, respectively, with hybrid success reaching 90.5\% and 87.4\% alongside reduced agentic API expenditure. On RoboTwin C2R, our method with 7.5K trajectories matches conventional training with 27.5K trajectories. Ablations validate reciprocal co-evolution, complementary solution routes in P2H, and counterfactual preference distillation in H2P. We further demonstrate autonomous task acquisition on MolmoSpaces~\cite{ai2molmospaces2026} and cross-embodiment harness transfer. On the Unitree G1-D, a simulation-evolved harness enables the collection of 120 demonstrations, yielding 85.0\% standalone success compared with 80.0\% using 200 human demonstrations, while fixed-harness hybrid execution reaches 95.0\%.

%% file: sections/2_related_work.tex
\section{Related Work}

\noindent \textbf{Continual Robot Learning.}
A continually learning policy must retain earlier skills
and obtain supervision for new ones.
For retention, EWC regularizes
weights~\cite{kirkpatrick2017ewc}, ER replays
data~\cite{rolnick2019er}, and PackNet isolates
parameters~\cite{mallya2018packnet}.
In manipulation, LOTUS builds skill
libraries~\cite{wan2024lotus} and TAIL uses task-specific
adapters~\cite{liu2024tail}.
On pretrained VLAs, experience replay mitigates
forgetting~\cite{liu2026vlaforgetting,zhu2026realworldcl};
PHASER refines it by prioritizing phases most at risk of
forgetting~\cite{chen2026phaser}, while CRL-VLA limits
advantage magnitudes on prior tasks during continual
RL~\cite{zeng2026crlvla}.
For new supervision, HG-DAgger and RECAP draw on human
interventions during
policy rollouts~\cite{kelly2019hgdagger,intelligence2025recap},
APO and FlowPRO turn such interventions into preference
signals~\cite{xia2025apo,wu2026flowpro}, and SOAR uses
autonomous VLM-guided practice~\cite{zhou2024soar}.
RoboEvo$^2$ learns from verified harness repairs of policy
failures rather than operator corrections, and employs failure-aware task sampling to retain skills
and improve low-success tasks.

\noindent \textbf{Agentic Self-Improvement.}
Agentic systems such as Code as
Policies~\cite{liang2023code}, Harness
VLA~\cite{zhang2026harness}, RATs~\cite{zhang2026rats},
and ASPIRE~\cite{aspire2026} solve tasks through programs,
skills, and memory without model fine-tuning.
Among recursive self-improvement (RSI) frameworks,
Zetta, EmbodiedEvo, EmbodiedRSI, and RPG evolve only the
harness~\cite{ding2026zetta,li2026embodiedevo,song2026embodiedrsi,wang2026rpg},
whereas ENPIRE and EmbodiRSI focus on policy
improvement~\cite{xiao2026enpire,lang2026embodirsi}.
Co-Harness and Self-HPO alternate harness and model
updates for LLM
agents~\cite{coharness2026,wei2026selfhpo}. AdaRSI~\cite{wang2026adarsi} refines its harness
and fine-tunes only sub-steps that still fail when retried
from saved states.
However, these methods train the policy only to imitate
successful executions.
In contrast, RoboEvo$^2$ also learns from failures: it pairs
a failed continuation with a verified repair from the same
state and trains the policy to prefer the repair.
Repairs are sought in parallel along code-only,
demonstration-guided, and policy-calling routes.

%% file: sections/3_method.tex
\begin{figure*}[t]
\centering
\includegraphics[width=\textwidth]{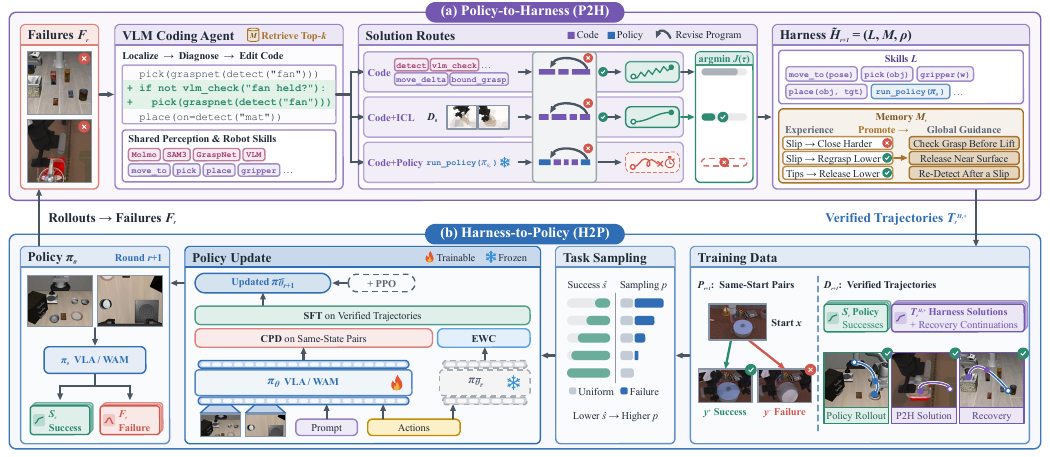}
\caption{\textbf{P2H and H2P evolution.}
(a)~P2H: for each failure in $\mathcal F_r$, the VLM coding agent retrieves memory,
diagnoses the failure, and searches Code, Code+ICL, and Code+Policy routes
in parallel; it keeps the verified solution with the lowest jerk $J(\tau)$
and merges compatible changes into the skills $L$, memory $M$, and routing
$\rho$ of $\widetilde{H}_{r+1}$.
(b)~H2P: verified trajectories and same-state success/failure pairs update
the policy with SFT, CPD, and EWC under failure-aware task sampling; the
updated policy starts round $r{+}1$.}
\label{fig:agentic-method}
\end{figure*}

\section{\texorpdfstring{RoboEvo$^2$}{RoboEvo2}}

\subsection{Overview of Recursive Co-Evolution}

RoboEvo$^2$ is a policy-driven framework for recursive
harness--policy co-evolution.
The current policy exposes capability gaps through
execution failures, the harness discovers effective
solutions, and the policy subsequently internalizes
these corrections.
As illustrated in \cref{fig:agentic-method},
RoboEvo$^2$ jointly evolves an agentic harness
$H_r=(L_r,M_r,\rho_r)$ and a neural policy
$\pi_{\theta_r}$ at evolution round $r$.

The harness consists of an executable skill library
$L_r$, persistent experience memory $M_r$, and a
routing mechanism $\rho_r$ for skill invocation
and code--policy coordination.
We initialize $H_0$ with robot APIs and parameterized
code-based skills from
CaP-X~\cite{fu2026capx} and
RATs~\cite{zhang2026rats}.
The initial policy $\pi_{\theta_0}$ is obtained
by fine-tuning a pretrained VLA or WAM on robot
demonstrations $\mathcal D_0$.

Each evolution round $r$ consists of four stages:

\noindent\textbf{(1) Policy Rollouts.}
The current policy $\pi_{\theta_r}$ serves as
the primary executor, generating task rollouts
$\mathcal T_r$.
Task-level verification partitions them into
successful trajectories $\mathcal S_r$ and
failed trajectories $\mathcal F_r$.
The failures identify where additional capabilities
are needed.

\noindent\textbf{(2) Policy-to-Harness Evolution (P2H).}
For each failed rollout, a VLM-based coding agent (GPT-6 Astra) diagnoses the failure and searches for solutions
through three complementary routes:
Code, Code+ICL, and Code+Policy.
Verified solutions are selected based on execution
quality, and compatible improvements are integrated
across failures into a candidate harness
$\widetilde H_{r+1}$.
Execution and repair experiences are retained
for subsequent reuse (Sec.~\ref{sec:p2h}).

\noindent\textbf{(3) Harness-to-Policy Evolution (H2P).}
Successful policy rollouts and harness-generated
corrections are accumulated as training data.
Matched successful and failed executions further
provide counterfactual supervision.
Task-balanced, failure-aware sampling and policy
optimization produce an updated candidate policy
$\pi_{\widetilde\theta_{r+1}}$
(Sec.~\ref{sec:h2p}).

\noindent\textbf{(4) Evaluation and Selection.}
We evaluate two execution modes:
\emph{standalone}, where the policy executes tasks
independently, and \emph{hybrid}, where the harness
coordinates policy execution with reusable code skills.
Candidate and previous versions are compared on
matched development tasks, initial states, and
execution budgets.
Updates are retained when they improve the corresponding
success rates; otherwise, previous versions are preserved.
The selected pair $(H_{r+1},\pi_{\theta_{r+1}})$
enters the next round.
Further details are provided in Appendix~\ref{sec:supp-benchmarks}.

By repeatedly discovering and internalizing
missing capabilities, RoboEvo$^2$ forms a recursive
improvement loop driven by the evolving policy's
remaining failures.

\subsection{Policy-to-Harness (P2H) Evolution}
\label{sec:p2h}

P2H evolution discovers executable solutions to
capability gaps exposed by the current policy.
Rather than relying on the coding agent to solve
tasks independently from scratch, P2H targets
failed policy executions and converts successful
corrections into reusable harness capabilities.
Given $\mathcal F_r$, it first searches for
solutions to individual failures and then
consolidates the verified results across tasks.

\noindent\textbf{Failure-Guided Solution Search and Selection.}
For each failed rollout $f\in\mathcal F_r$,
a VLM-based coding agent diagnoses the failure
using policy observations, action history,
and task outcomes.
It retrieves relevant skills from $L_r$ and
past experiences from $M_r$, then explores
three complementary solution routes in parallel:

\begin{itemize}
    \item \textbf{Code:}
    The agent first attempts recovery using the
    code-based skills in the current harness.
    If unsuccessful, it refines executable skills,
    including motion targets, grasp and contact
    strategies, phase transitions, and recovery logic
    based on execution feedback.

    \item \textbf{Code+ICL:}
    The agent randomly samples one demonstration
    for the target task from $\mathcal D_0$
    to guide harness refinement.
    It extracts task structure, manipulation
    strategies, and temporal cues to revise
    executable skills, including object-relative
    motions, arm coordination, gripper control,
    and execution sequencing.
    The revisions are then tested and
    iteratively improved via execution feedback.

    \item \textbf{Code+Policy:}
    The agent first attempts recovery by coordinating
    code-based skills in the current harness with
    the policy $\pi_{\theta_r}$ as a learned
    motor primitive.
    If unsuccessful, it refines harness orchestration,
    including subgoal allocation, skill composition,
    policy invocation, handoff conditions, and
    recovery strategies.
    This route exploits the complementary capabilities
    of code and learned policies.
\end{itemize}

The three routes explore different solution
strategies while sharing the same task goal
and execution constraints.
Each route operates in an isolated execution branch,
starting from the relevant failure state when feasible
or from a matched task reset otherwise.
Execution and refinement continue until task success
or budget exhaustion, while the policy parameters
remain frozen throughout P2H.

Each route retains at most one verified successful
program--trajectory pair for failure $f$, forming
a candidate set $\mathcal C_f^+$ of up to three
solutions.
Among successful candidates evaluated under
matched conditions, we select the one with the
lowest normalized motion jerk $J(\tau)$~\cite{li2024open}:
\begin{equation}
C_f^*=(c_f^*,\tau_f^*)
\in \operatorname*{arg\,min}_{(c,\tau)\in\mathcal C_f^+}
J(\tau),
\end{equation}
where $c_f^*$ and $\tau_f^*$ denote the selected
executable program and verified trajectory,
respectively.
If no candidate succeeds, the failure remains
unresolved and is retained for diagnosis.

Importantly, each resolved failure associates
the original failed policy execution with a
verified successful correction.
These execution records provide targeted
supervision for subsequent policy learning,
including matched corrective pairs for CPD
when the corresponding state and execution
conditions are satisfied (Sec.~\ref{sec:h2p}).
Further implementation details are provided
in Appendices~\ref{sec:supp-harness} and~\ref{sec:supp-training}.

\noindent\textbf{Cross-Failure Harness Integration.}
Beyond individual recovery, P2H consolidates
successful solutions into persistent capabilities
that can be reused across tasks.
The coding agent merges compatible modifications
from selected solutions into a unified candidate
harness $\widetilde H_{r+1}$.

The skill library $L_r$ is extended with new or
refined executable programs, while the routing
mechanism $\rho_r$ is updated when skill selection,
subgoal coordination, or code--policy switching
changes.
Meanwhile, memory $M_r$ accumulates failure
contexts, attempted solutions, execution outcomes,
and validated repair guidance.
Successful recovery does not necessarily require
modifying code; solutions using existing skills
still contribute verified trajectories and
reusable experience.
The integrated harness is checked for compatibility
and evaluated before replacing $H_r$, preventing
individual repairs from automatically becoming
persistent system updates. Further details are provided in Appendix~\ref{sec:supp-harness}.

\subsection{Harness-to-Policy (H2P) Evolution}
\label{sec:h2p}

H2P evolution aims to internalize agent-discovered
corrective behaviors into the neural policy.
It leverages two complementary forms of supervision.
First, successful policy rollouts $\mathcal S_r$
and verified harness-generated trajectories
$\mathcal T_r^{H,+}$, including full-task solutions
and recovery continuations, are accumulated in
$\mathcal D_{r+1}$ for supervised learning.
Second, matched policy failures and successful
corrections form a corrective preference dataset
$\mathcal P_{r+1}$ for Counterfactual Preference
Distillation (CPD), which encourages the policy
to favor corrected behaviors over failed alternatives.

\noindent\textbf{Counterfactual Preference Distillation.}
We build $\mathcal P_{r+1}$ by pairing original
policy failures with verified harness-generated
corrections.
For each resolved failure, successful and failed
continuations are compared from the same starting
state under matched execution conditions.
Each valid pair is represented as $(x,y^+,y^-)$,
where $x=(o,g)$ denotes the shared observation
and task goal, and $y^+$ and $y^-$ denote
successful and failed action targets, respectively.
Only outcome-verified, state-matched pairs enter
$\mathcal P_{r+1}$; unmatched successful recoveries
remain available in $\mathcal D_{r+1}$.

To learn from these corrective pairs, CPD optimizes
the relative losses of successful and failed targets
against a frozen reference policy.
Let $\ell_\theta(y\mid x)$ denote the per-sample
training loss, and let $\bar\theta_r=\theta_r$
denote the frozen policy parameters at the
beginning of round $r$.
We define
\begin{equation}
\delta^\pm=\ell_\theta(y^\pm\mid x)-\ell_{\bar\theta_r}(y^\pm\mid x),
\end{equation}
with $\Delta=\delta^+-\delta^-$ measuring the
change in the successful-versus-failed loss gap
relative to the reference policy.

CPD combines a margin-based preference objective
with asymmetric proximity constraints:
\begin{equation}
\begin{aligned}
\mathcal L_{\mathrm{gap}}
&=\mathbb E\left[
\bigl(\max\{\Delta,-m\}+m\bigr)^2
\right],\\
\mathcal L_{\mathrm{prox}}
&=\mathbb E\left[
[\delta^+]_+^2+
[\delta^--\kappa]_+^2
\right],
\end{aligned}
\end{equation}
where $[z]_+=\max(z,0)$,
$m$ is the preference margin,
and $\kappa$ controls the allowed increase in
negative-target loss.
The gap term favors successful corrections
over failed alternatives, while the asymmetric
proximity term prevents degradation on successful
targets and excessive suppression of failed ones.
Together, they promote corrective behavior
learning while maintaining proximity to $\bar\theta_r$.

\noindent\textbf{Failure-Aware Continual Policy Optimization.}
To prioritize current capability gaps while
mitigating forgetting, we adopt task-balanced,
failure-aware sampling for both successful
trajectories $\mathcal D_{r+1}$ and corrective
pairs $\mathcal P_{r+1}$.
Let $\mathcal K_b$ denote the tasks represented
in data pool $b\in\{\mathcal D,\mathcal P\}$,
and $\hat s_{r,k}$ the current policy's success
rate on task $k$.
With failure weight
$w_{r,k}=(1-\hat s_{r,k})^\gamma$, we define
\begin{equation}
p_{r,k}^{(b)}=\frac{1}{2|\mathcal K_b|}+\frac{w_{r,k}}{2\sum_{j\in\mathcal K_b}w_{r,j}},
\end{equation}
where $\gamma>0$ controls the emphasis on
challenging tasks.
We independently sample supervised trajectories
and corrective pairs from their respective
task distributions, balancing historical
coverage with targeted improvement.
Each policy uses its native supervised
objective $\mathcal L_{\mathrm{sup}}$.
With compatible corrective
pairs, the update objective is
\begin{equation}
\mathcal L_{\mathrm{update}}=\mathcal L_{\mathrm{sup}}+\lambda_{\mathrm{pref}}(\mathcal L_{\mathrm{gap}}+ \lambda_{\mathrm{prox}}\mathcal L_{\mathrm{prox}}).
\end{equation}
Without compatible pairs, we set
$\lambda_{\mathrm{pref}}=0$.
To further mitigate forgetting, we incorporate
Elastic Weight Consolidation (EWC)~\cite{kirkpatrick2017ewc},
which penalizes changes to parameters important
for previously learned tasks:
\begin{equation}
\mathcal L_{\mathrm{total}}=
\mathcal L_{\mathrm{update}}+
\frac{\lambda_{\mathrm{EWC}}}{2}
\sum_i F_{r,i}(\theta_i-\bar\theta_{r,i})^2.
\end{equation}
Here, $i$ indexes the policy parameters,
$\lambda_{\mathrm{EWC}}$ controls the regularization
strength, and $F_{r,i}$ estimates the importance
of parameter $i$ using squared per-example gradients.

After offline co-evolution, compatible action
policies can be further refined using
PPO~\cite{schulman2017ppo}, which collects fresh
simulator rollouts and optimizes task-success rewards.

\subsection{Standalone and Hybrid Execution Modes}

RoboEvo$^2$ supports two complementary execution
modes that balance policy autonomy and agentic
assistance:
\begin{itemize}
    \item \textbf{Standalone Execution:}
    The co-evolved policy executes tasks directly
    without coding-agent intervention, enabling
    efficient neural inference while leveraging
    capabilities acquired from agentic corrections.

    \item \textbf{Hybrid Execution:}
    An embodied agent coordinates the co-evolved
    policy with reusable code skills, supporting
    explicit task decomposition, skill composition,
    and feedback-driven recovery~\cite{zhang2026harness}.
\end{itemize}

Through recursive co-evolution,
policy failures drive the discovery of new
executable capabilities, while agentic corrections
are progressively internalized into the policy.
As the policy improves, its evolving capability gaps
guide subsequent harness evolution, forming a
self-reinforcing cycle of discovery and learning.
This process aims to amortize costly agentic reasoning
into efficient policy execution, progressively reducing
reliance on agentic assistance while retaining the
harness for unresolved challenges.

%% file: sections/4_experiments.tex
\section{Experiments}

We evaluate RoboEvo$^2$ on LIBERO-PRO~\cite{zhou2025liberopro}
and RoboTwin C2R~\cite{chen2025robotwin2} for
multi-round co-evolution and generalization to unseen
perturbations, with ablations examining reciprocal
evolution, P2H solution routes, and CPD.
MolmoSpaces~\cite{ai2molmospaces2026} evaluates
autonomous capability acquisition, while cross-embodiment
and Unitree G1-D experiments demonstrate harness
reusability, sim-to-real transfer, and real-world
policy learning.

\subsection{Generalization and Multi-Round Evolution}

We evaluate RoboEvo$^2$ on two benchmarks under distribution shifts. LIBERO-PRO~\cite{zhou2025liberopro} transfers policies trained on original LIBERO demonstrations to task and position perturbations, while RoboTwin C2R~\cite{chen2025robotwin2} adapts policies trained on 2,500 clean demonstrations to randomized scenes across 50 bimanual tasks. On both benchmarks, evolution runs on seeds held out from the initial training data without additional human annotation, and generalization is evaluated on separate unseen seeds (800 LIBERO-PRO and 5,000 RoboTwin C2R episodes; Appendix~\ref{sec:supp-benchmarks}).

\begin{figure}[t]
\centering
\captionsetup{skip=4pt}
\includegraphics[width=0.95\linewidth]
{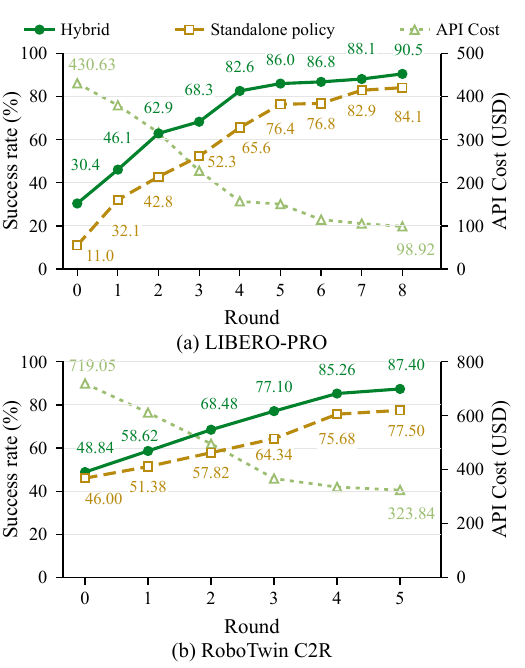}
\caption{\textbf{Multi-round evolution.}
Standalone-policy and hybrid success rates
improve on (a) LIBERO-PRO and (b) RoboTwin C2R,
while per-round coding-agent API cost decreases.}
\label{fig:coevolution-rounds}
\vspace{-3pt}
\end{figure}

\noindent\textbf{Multi-Round Adaptation and Data Efficiency.}
\Cref{fig:coevolution-rounds} demonstrates progressive improvements in standalone and hybrid execution. On LIBERO-PRO, eight rounds improve standalone success from 11.0\% to 84.1\% and hybrid success from 30.4\% to 90.5\%, while API expenditure decreases from \$430.63 to \$98.92. On RoboTwin C2R, five rounds increase standalone success from 46.0\% to 77.5\% and hybrid success from 48.8\% to 87.4\%, with API expenditure decreasing from \$719.05 to \$323.84. These results demonstrate sustained capability improvement alongside decreasing agentic costs, highlighting the benefits of recursive harness--policy co-evolution.

Beyond execution efficiency, RoboEvo$^2$ acquires training experience without additional human demonstrations. As shown in \cref{tab:robotwin-c2r}, starting from 2,500 clean demonstrations, our framework autonomously collects 5,000 verified trajectories and achieves 77.5\% standalone success on RoboTwin C2R. With the same 7.5K trajectories, clean+random training reaches only 67.2\%, and it needs 27.5K to reach 76.8\%. RoboEvo$^2$ thus matches it with only 27.3\% of the training data, suggesting that failure-driven experience provides effective supervision.

\noindent\textbf{Comparison with Self-Evolving Methods.}
\Cref{tab:liberopro-main} compares RoboEvo$^2$ with recent self-evolving robot systems. Harness-centered methods, including ASPIRE~\cite{aspire2026}, Zetta~\cite{ding2026zetta}, DynaHarness~\cite{deng2026dynaharness}, and EmbodiedRSI~\cite{song2026embodiedrsi}, primarily improve executable skills and agentic recovery. In contrast, RoboEvo$^2$ enables reciprocal evolution, where harness discoveries provide corrective supervision for policy learning, while the updated policy supplies stronger motor primitives and guides subsequent harness search. Across all eight LIBERO-PRO settings, RoboEvo$^2$ achieves 84.1\% standalone and 90.5\% hybrid success, surpassing EmbodiedRSI's 86.8\% hybrid success. Further PPO refinement increases standalone and hybrid success to 94.8\% and 96.8\%, respectively. These results demonstrate competitive adaptation performance while progressively internalizing agent-discovered capabilities into the policy.

\begin{table}[t]
\centering
\caption{\textbf{Data efficiency on RoboTwin C2R.}
Success rates (\%) with different data.
Rand denotes randomized demonstrations, while
Auto denotes autonomously collected trajectories.}
\label{tab:robotwin-c2r}
\small
\setlength{\tabcolsep}{2.8pt}
\begin{tabular}{l|cccc|c}
\toprule
Method & Clean & Rand & Auto & Total & SR (\%) \\
\midrule
$\pi_{0.5}$ (clean)
& 2.5K & -- & -- & 2.5K & 46.0 \\
$\pi_{0.5}$ (+5K rand)
& 2.5K & 5K & -- & 7.5K & 67.2 \\
$\pi_{0.5}$ (+25K rand)
& 2.5K & 25K & -- & 27.5K & 76.8 \\
\midrule
\textbf{RoboEvo$^2$} ($\pi_{0.5}$)
& 2.5K & -- & 5K & 7.5K & \textbf{77.5} \\
\textbf{RoboEvo$^2$} (Hybrid)
& 2.5K & -- & 5K & 7.5K & \textbf{87.4} \\
\bottomrule
\end{tabular}
\end{table}

\begin{table*}[t]
\centering
\caption{\textbf{Comparison with self-evolving methods on LIBERO-PRO.}
Success rates (\%). Methods are grouped by their primary evolution mechanisms.
Avg.$_4$: Goal and LIBERO-10; Avg.$_8$: all eight settings.
Published results retain their original evaluation protocols.}
\label{tab:liberopro-main}
\small
\setlength{\tabcolsep}{4pt}
\resizebox{\textwidth}{!}{
\begin{tabular}{l|l|cccccccc|cc}
\toprule
Category & Method & Spat-T & Spat-S & Obj-T & Obj-S
& Goal-T & Goal-S & L10-T & L10-S
& Avg.$_4$ & Avg.$_8$ \\
\midrule

\multirow{4}{*}{\shortstack[l]{Harness\\Evolution}}
& ASPIRE~\cite{aspire2026}
& 60.0 & 51.0 & 95.0 & 98.0
& 45.0 & 81.0 & 38.3 & 22.6
& 46.7 & 61.4 \\

& Zetta~\cite{ding2026zetta}
& -- & -- & -- & --
& 92.5 & 89.0 & 63.0 & 40.0
& 71.1 & -- \\

& DynaHarness~\cite{deng2026dynaharness}
& -- & -- & -- & --
& 79.0 & 78.0 & 77.5 & 66.5
& 75.3 & -- \\

& EmbodiedRSI~\cite{song2026embodiedrsi}
& \textbf{100.0} & 93.0 & \textbf{100.0} & 96.0
& 78.0 & 90.0 & 66.0 & 71.0
& 76.3 & 86.8 \\

\midrule
\multirow{6}{*}{\shortstack[l]{Reciprocal\\Co-Evolution}}
& \textbf{RoboEvo$^2$} (FastWAM, standalone)
& 81.0 & 60.0 & \textbf{100.0} & \textbf{100.0}
& 75.0 & 73.0 & 66.0 & 49.0
& 65.8 & 75.5 \\

& \textbf{RoboEvo$^2$} (FastWAM, hybrid)
& 95.0 & 73.0 & \textbf{100.0} & \textbf{100.0}
& 85.0 & 83.0 & 73.0 & 54.0
& 73.8 & 82.9 \\

& \textbf{RoboEvo$^2$} ($\pi_{0.5}$, standalone)
& 93.0 & 93.0 & 98.0 & 99.0
& 84.0 & 82.0 & 65.0 & 59.0
& 72.5 & 84.1 \\

& \textbf{RoboEvo$^2$} ($\pi_{0.5}$, hybrid)
& 97.0 & 92.0 & 98.0 & \textbf{100.0}
& \textbf{96.0} & 95.0 & 72.0 & 74.0
& 84.3 & 90.5 \\

& \textbf{RoboEvo$^2$} ($\pi_{0.5}$ + PPO, standalone)
& 97.0 & 98.0 & 98.0 & \textbf{100.0}
& 90.0 & 98.0 & 97.0 & 80.0
& 91.3 & 94.8 \\

& \textbf{RoboEvo$^2$} ($\pi_{0.5}$ + PPO, hybrid)
& \textbf{100.0} & \textbf{99.0} & 98.0 & \textbf{100.0}
& 95.0 & \textbf{99.0} & \textbf{98.0} & \textbf{85.0}
& \textbf{94.3} & \textbf{96.8} \\
\bottomrule
\end{tabular}
}
\end{table*}

\subsection{Ablation Studies}

We conduct ablation studies on LIBERO-PRO to evaluate
the contributions of individual components in RoboEvo$^2$.

\noindent\textbf{Effect of Joint Harness--Policy Evolution.}
We compare full RoboEvo$^2$ with two variants that freeze either the policy (P2H only) or the harness (H2P only).
As shown in \cref{tab:ablation-rsi}, full co-evolution consistently outperforms both variants, reaching 52.3\% standalone and 68.3\% hybrid success after three rounds.
Notably, its standalone advantage over H2P only increases across rounds, suggesting that evolving harness capabilities provide increasingly effective supervision for policy learning.
These results highlight the benefits of reciprocal capability discovery and internalization over one-sided evolution.

\begin{table}[t]
\centering
\caption{\textbf{Effect of joint harness--policy evolution
on LIBERO-PRO.} Standalone policy (P) and hybrid (H)
success rates (\%) across three evolution rounds.}
\label{tab:ablation-rsi}
\small
\resizebox{\columnwidth}{!}{
\begin{tabular}{l|cc|cc|cc}
\toprule
& \multicolumn{2}{c|}{Round 1}
& \multicolumn{2}{c|}{Round 2}
& \multicolumn{2}{c}{Round 3} \\
\cline{2-7}
Setting & P & H & P & H & P & H \\
\midrule
P2H only (fixed policy)
& -- & 35.9 & -- & 42.6 & -- & 46.1 \\
H2P only (fixed harness)
& 22.3 & 37.0 & 28.5 & 44.8 & 33.6 & 49.0 \\
\midrule
\textbf{Full RoboEvo$^2$}
& 32.1 & \textbf{46.1}
& \textbf{42.8} & \textbf{62.9}
& \textbf{52.3} & \textbf{68.3} \\
\bottomrule
\end{tabular}
}
\end{table}

\noindent\textbf{Effect of Complementary P2H Solution Routes.}
Starting from the same policy rollouts with an initial
standalone success rate of 11.0\%, we evaluate
Code, Code+ICL, Code+Policy, and their combination
in the first evolution round.
Each route first attempts execution using the existing
harness without modification, then iteratively repairs
failed executions through feedback-driven refinement.
As shown in \cref{tab:ablation-p2h}, all routes recover
additional failures, with Code+Policy achieving the
largest individual gain of 14.8 points.
Combining all routes achieves the highest success rate
of 49.1\%, highlighting their complementary
problem-solving capabilities.
Training on the collected trajectories further
improves standalone policy success from 11.0\%
to 32.1\% with all routes, showing that
complementary discovery supports
policy internalization.

\begin{table}[t]
\centering
\caption{\textbf{Effect of P2H solution routes on LIBERO-PRO.}
First-round task success rates (\%) with and without
repair, net recovery gain
in percentage points (pp), and standalone policy success
after training on the collected trajectories.
The initial policy achieves 11.0\% success.}
\label{tab:ablation-p2h}
\small
\resizebox{\columnwidth}{!}{
\begin{tabular}{l|ccc|c}
\toprule
Solution Route & w/o Repair & w/ Repair & Gain (pp) & Policy SR \\
\midrule
Code & 21.6 & 29.4 & +7.8 & 21.0 \\
Code+ICL & 31.5 & 44.9 & +13.4 & 28.6 \\
Code+Policy & 30.8 & 45.6 & +14.8 & 29.0 \\
\midrule
\textbf{All Routes} & 35.6 & \textbf{49.1} & +13.5 & \textbf{32.1} \\
\bottomrule
\end{tabular}
}
\end{table}

\noindent\textbf{Effect of CPD Loss Components.}
Through P2H evolution, multi-route agentic recovery
produces successful corrections to policy failures,
providing matched corrective pairs for H2P learning.
We evaluate the preference gap
$\mathcal L_{\mathrm{gap}}$ and asymmetric proximity
term $\mathcal L_{\mathrm{prox}}$ using these pairs
under identical training data, initialization,
and EWC settings.
As shown in \cref{tab:ablation-cpd}, adding the
preference gap to supervised learning improves
standalone success from 20.1\% to 29.9\%.
The full CPD objective further increases success
to 32.1\%, demonstrating the benefit of asymmetric
proximity constraints for corrective capability
internalization.

\begin{table}[t]
\centering
\caption{\textbf{Effect of CPD loss components on LIBERO-PRO.}
}
\label{tab:ablation-cpd}
\small
\resizebox{\columnwidth}{!}{
\begin{tabular}{l|ccc}
\toprule
Objective
& $\mathcal L_{\mathrm{sup}}$
& $+\lambda_{\mathrm{pref}}\mathcal L_{\mathrm{gap}}$
& $+\lambda_{\mathrm{pref}}(\mathcal L_{\mathrm{gap}}+\lambda_{\mathrm{prox}}\mathcal L_{\mathrm{prox}})$ \\
\midrule
SR (\%) & 20.1 & 29.9 & \textbf{32.1} \\
\bottomrule
\end{tabular}
}
\end{table}

\begin{table}[t]
\centering
\caption{\textbf{Task acquisition on MolmoSpaces.} Success rates (\%).}
\label{tab:molmospaces-benchmark}
\small
\setlength{\tabcolsep}{3pt}
\begin{tabular}{l|cccc|c}
\toprule
Method & Open & Close & Pick & Pick\&Place & Avg. \\
\midrule
CaP-Agent0~\cite{fu2026capx} & 14 & 36 & 23 & 11 & 21.0 \\
RATs~\cite{zhang2026rats} & 20 & 73 & 37 & 22 & 38.0 \\
\midrule
\textbf{RoboEvo$^2$} (Policy)
& 9 & 99 & 70 & 50 & 57.0 \\
\textbf{RoboEvo$^2$} (Hybrid)
& \textbf{32} & \textbf{100} & \textbf{73}
& \textbf{58} & \textbf{65.8} \\
\bottomrule
\end{tabular}
\end{table}

\subsection{Capability Acquisition and Transfer}

We further investigate whether RoboEvo$^2$ can
autonomously acquire manipulation capabilities
and reuse evolved skills across robot embodiments.

\noindent\textbf{Autonomous Task Acquisition.}
MolmoSpaces~\cite{ai2molmospaces2026} evaluates
autonomous manipulation skill acquisition, where
agents propose scene-grounded goals and discover
executable solutions through exploration.
Unlike code-only agents such as CaP-Agent0~\cite{fu2026capx} and
RATs~\cite{zhang2026rats}, we extend the exploration
process of RATs by incorporating a trainable policy
within our harness--policy co-evolution framework.
Verified agentic executions are used to improve
the policy, enabling acquired skills to be
internalized beyond executable programs.
Across 40 tasks spanning opening, closing, picking,
and pick-and-place, our standalone policy achieves
57.0\% success, compared with 38.0\% for RATs,
while hybrid execution further reaches 65.8\%
(\cref{tab:molmospaces-benchmark}).

\noindent\textbf{Cross-Embodiment Harness Transfer.}
We evaluate whether the harness evolved on RoboTwin
with ALOHA AgileX can be directly reused across
robot embodiments.
Keeping the same 50 tasks, 20 held-out layouts per task, executable
skills, routing logic, and persistent memory,
we replace the robot with dual ARX-X5 or
dual Franka-Panda, adapting only the robot-specific
control interfaces.
As shown in \cref{tab:embodiment-transfer},
the frozen harness achieves 45.8\% success on
ALOHA AgileX, 42.5\% on dual ARX-X5, and 41.0\%
on dual Franka-Panda.
The small performance drops indicate that
evolved harness capabilities can transfer
across robot embodiments without additional
skill evolution.

\begin{table}[t]
\centering
\caption{\textbf{Cross-embodiment harness transfer.}
The frozen harness is evaluated on the same
RoboTwin tasks and layouts.}
\label{tab:embodiment-transfer}
\small
\begin{tabular}{l|ccc}
\toprule
Robot & ALOHA AgileX & ARX-X5 & Franka-Panda \\
\hline
SR (\%) & \textbf{45.8} & 42.5 & 41.0 \\
\bottomrule
\end{tabular}
\end{table}

\begin{table}[t]
\centering
\caption{\textbf{Real-world manipulation on the Unitree G1-D.}
Task success rates (\%) over 10 trials per task.
Teleop denotes human demonstrations; Auto denotes
automatically collected demonstrations.
The harness is transferred from RoboTwin.}
\label{tab:G1-D-results}
\small
\resizebox{\columnwidth}{!}{
\begin{tabular}{l|cccc|c}
\toprule
Method & Stack & Relocate & Insert & Retrieve & Avg. \\
\midrule
$\pi_{0.5}$ (200 Teleop)
& 80 & 90 & 60 & 90 & 80.0 \\
Fixed Harness (Code)
& 40 & 70 & 30 & 50 & 47.5 \\
\midrule
\textbf{RoboEvo$^2$} (120 Auto)
& 90 & 90 & 70 & 90 & 85.0 \\
\textbf{RoboEvo$^2$} (Hybrid)
& \textbf{100} & \textbf{100} & \textbf{80}
& \textbf{100} & \textbf{95.0} \\
\bottomrule
\end{tabular}
}
\end{table}

\subsection{Real-World Robot Manipulation}

\begin{figure}[t]
\centering
\includegraphics[width=\linewidth]
{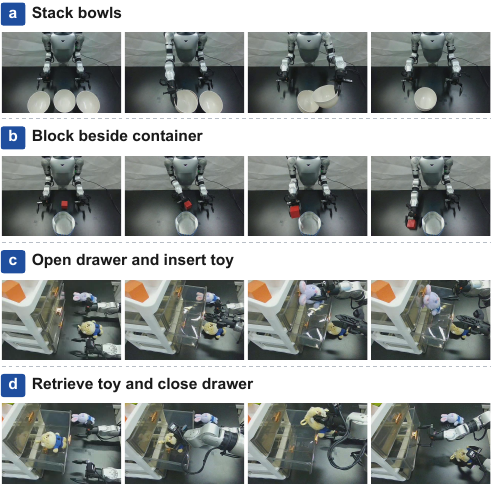}
\caption{\textbf{Real-world executions on the Unitree G1-D.}
}
\label{fig:G1-D-executions}
\end{figure}

We evaluate RoboEvo$^2$ on four bimanual tasks (\cref{fig:G1-D-executions})
using the Unitree G1-D, investigating whether
simulation-evolved harness capabilities can transfer
to physical robots and provide training supervision
without teleoperation.

\noindent\textbf{Harness Transfer and Autonomous Data Collection.}
We directly transfer the harness evolved on RoboTwin
to the Unitree G1-D, adapting only the robot interfaces
and estimating motion targets from real observations.
Using its frozen Code route, the transferred harness
collects 10 verified demonstrations per task.
We first fine-tune $\pi_{0.5}$ on these trajectories
and execute the resulting policy on the physical robot.
When execution fails, an operator approximately
restores the robot and scene to the pre-failure state,
without teleoperating the corrective actions.
The agent then explores Code, Code+ICL, and
Code+Policy routes, selecting verified corrections
with the lowest normalized motion jerk.
This process collects 20 additional successful
trajectories per task, yielding 120 automatically
generated demonstrations in total.
We use these trajectories and eligible corrective
pairs to train the final standalone policy through H2P.

\noindent\textbf{Real-World Results.}
As shown in \cref{tab:G1-D-results},
the transferred harness achieves 47.5\% mean success
without target-side skill evolution.
The RoboEvo$^2$ policy trained on 120 automatically
collected demonstrations achieves 85.0\% standalone
success, compared with 80.0\% for a $\pi_{0.5}$
baseline trained on 200 human-teleoperated demonstrations.
Combining the learned policy with the fixed harness
further increases success to 95.0\%.
These results demonstrate that reusable harness
capabilities can support sim-to-real execution
and generate effective policy supervision with
limited human intervention, while hybrid execution
continues to complement the learned policy.

%% file: sections/5_conclusion.tex
\section{Conclusion}

We presented RoboEvo$^2$, a framework for failure-driven recursive harness--policy co-evolution. Through complementary solution discovery (P2H) and counterfactual capability internalization (H2P), verified agentic corrections become reusable harness skills and learned policy behaviors. Across rounds, the improved policy provides stronger motor primitives and redirects subsequent harness exploration, closing the reciprocal learning loop. Experiments demonstrate progressive improvements in standalone and hybrid execution with reduced agentic costs, alongside data-efficient adaptation, autonomous task acquisition, cross-embodiment reuse, and real-world policy learning. RoboEvo$^2$ thus supports progressive policy autonomy while retaining agentic capabilities for unresolved challenges.

%% file: sections/6_llm_usage_statement.tex
\section*{LLM Usage Statement}
An LLM-based assistant (OpenAI Codex) was used for language
editing, LaTeX formatting, and checking
bibliographic metadata. All assisted content was reviewed and verified by the
authors. Separately, as part of the method, RoboEvo$^2$ uses GPT-6 Astra as its
coding agent (Qwen3.8-27B in the agent-backbone ablation; see
Appendix~\ref{sec:supp-agent-ablation}). No LLM is listed as an author; the authors take full responsibility
for the manuscript, implementation, experiments, and scientific claims.

%% file: sections/A_supplement.tex
\definecolor{suppnavy}{HTML}{1F3A6E}\definecolor{suppgray}{HTML}{7A7A7A}
\newcommand{\rlab}[1]{\textcolor{suppnavy}{\textbf{#1}}\enspace}
\newcommand{\rentry}[1]{\par\hangindent=1em\hangafter=1\noindent #1}
\newcommand{\repaircase}[7]{\noindent\begin{minipage}[t]{0.36\textwidth}\vspace{0pt}#1\end{minipage}\hfill
\begin{minipage}[t]{0.61\textwidth}\vspace{0pt}\footnotesize\raggedright
{\small\textbf{#2}}\enspace\textcolor{suppgray}{\textit{#3}}%
\rentry{\rlab{When}#4}\rentry{\rlab{Change}#5}\rentry{\rlab{Edits}#6}\rentry{\rlab{Case}#7}
\end{minipage}\par\vspace{4pt}}
\lstdefinestyle{promptstyle}{basicstyle=\footnotesize\ttfamily,columns=fullflexible,keepspaces=true,breaklines=true,breakatwhitespace=true,breakindent=0pt,backgroundcolor=\color{black!5},frame=single,framerule=0pt,framesep=4pt,xleftmargin=4pt,xrightmargin=4pt,aboveskip=8pt,belowskip=8pt,upquote=true,showstringspaces=false}
\newcommand{\promptbegin}{\par}
\lstnewenvironment{promptblock}{\promptbegin\lstset{style=promptstyle}}{}

\section{Harness Details}
\label{sec:supp-harness}

\subsection{Agent, Skill Library, Memory, and Routing}
\label{sec:supp-skills}

\paragraph{Coding agent.}
The agent is GPT-6 Astra (\texttt{gpt-6-astra}), run with reasoning effort \texttt{xhigh}; it is used both for coding and for demonstration video analysis. Each coding call receives the task instruction, recorded camera frames, the localized failure, retrieved memory entries, and the editable skill programs, and returns a program revision; the prompts are given in \cref{sec:supp-prompt}. After each round we save the accepted code, memory, policy checkpoint, and data manifest.

\paragraph{Skill library.}
The skill library $L_r$ holds only executable code, separate from memory $M_r$ and routing $\rho_r$. It is organized as a two-level skill tree. At the bottom are \emph{atomic skills}: observing, gripper commands, relative motions, motions to a target anchored at an image pixel, and policy calls, which only Code+Policy may use. Above them are \emph{composite skills}, each composed from atomic skills and parameterized by its arguments, such as grasping, placing, transferring, pulling, and stroking along a surface. A program calls skills by name with arguments. Extending $L_r$ adds a new entry, and refining a skill writes a new revision of its entry while earlier revisions are kept. \Cref{fig:supp-evolution} shows how general skills are specialized and how memory grows on LIBERO-PRO.

\paragraph{Memory.}
Memory $M_r$ has two parts. General memory holds lessons that apply across tasks, grouped by skill category (perception, grasping, placing, contact, policy, and motion and planning). Experiences are learned from individual tasks: each entry states when it applies, the action to take, and the skills it uses, and is retrieved by matching the task instruction. An experience that holds across tasks is promoted to general memory (\cref{fig:supp-evolution}).

\begin{figure}[t]
\centering
\includegraphics[width=\linewidth]{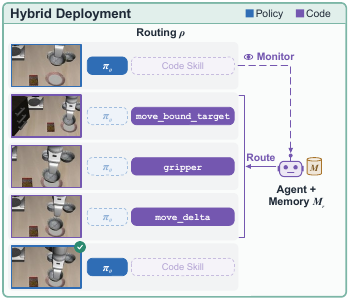}
\caption{\textbf{Hybrid deployment.} The agent with memory $M_r$ routes each step of a recorded LIBERO-PRO episode to the policy $\pi_\theta$ or a code skill (routing $\rho$).}
\label{fig:supp-hybrid}
\end{figure}

\paragraph{Routing.}
Routing $\rho_r$ is a set of rules in the agent's instructions rather than a separate model. In hybrid execution, code staging precedes each policy call, and after the call the agent re-observes before choosing the next step. Updating $\rho_r$ edits these rules. \Cref{fig:supp-hybrid} shows the routing on one recorded LIBERO-PRO episode.

\begin{figure*}[t]
\centering
\includegraphics[width=0.925\textwidth]{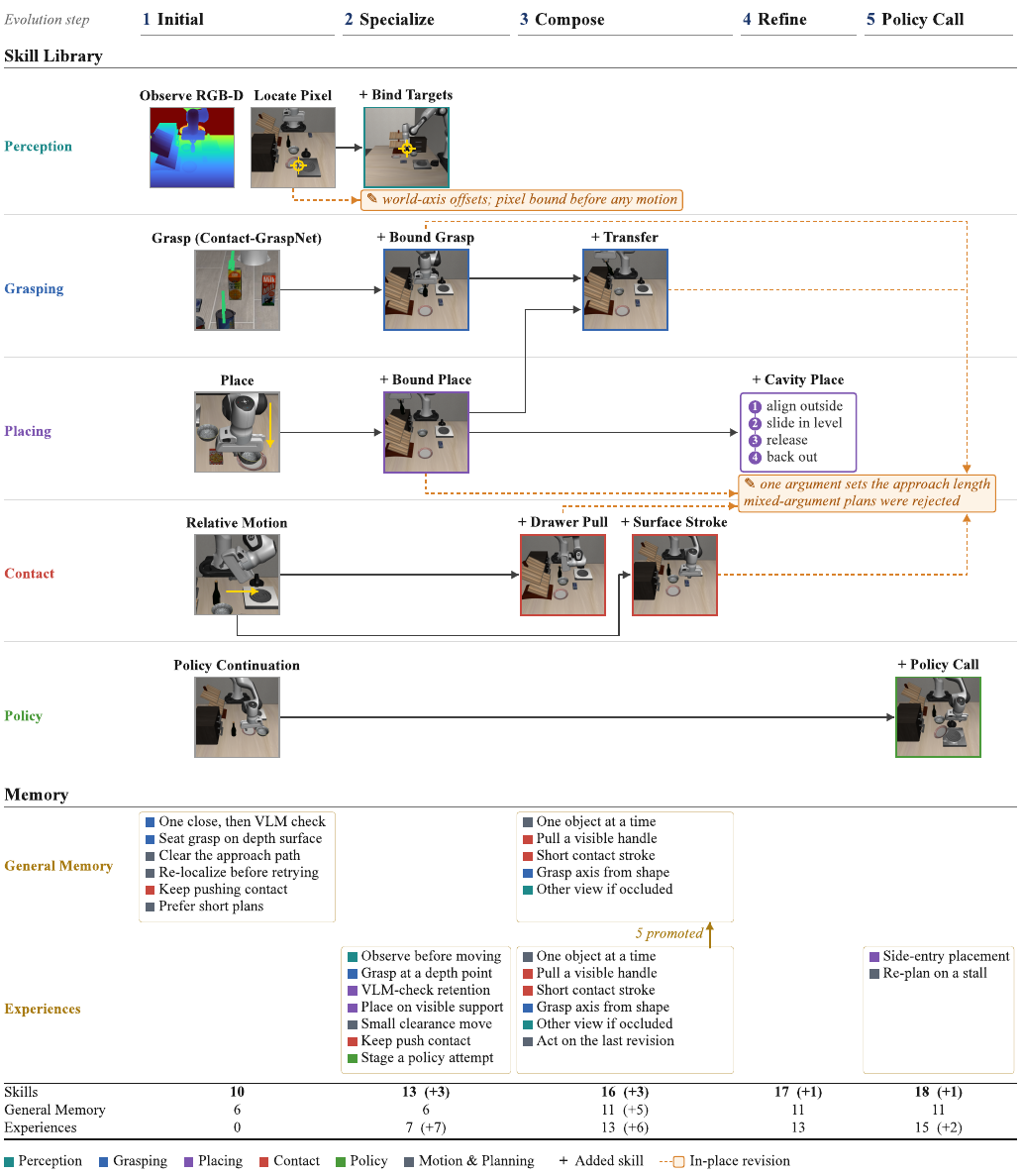}
\caption{\textbf{Harness evolution on LIBERO-PRO.} An example showing part of the harness over five evolution steps.}
\label{fig:supp-evolution}
\end{figure*}

\subsection{Solution Routes}

\begin{figure*}[t]
\centering
\includegraphics[width=\textwidth]{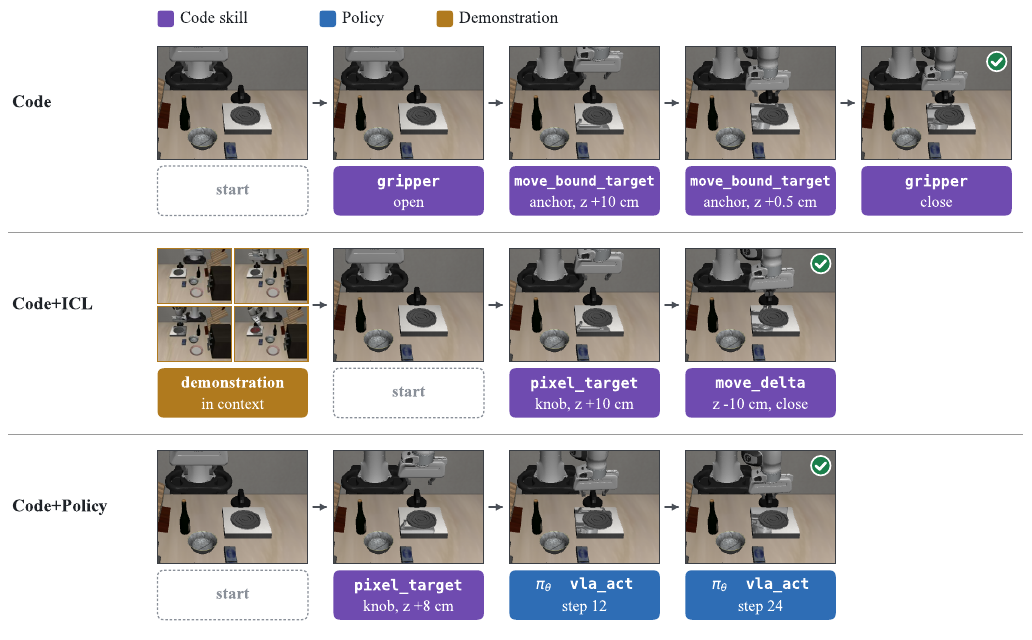}
\caption{\textbf{Solution routes on one LIBERO-PRO case} (\emph{turn off the stove}). Each panel is the agent view after the call below it; purple marks code skills, blue policy calls, and ochre the Code+ICL demonstration.}
\label{fig:libero-routes}
\end{figure*}

\paragraph{Route capabilities.}
Each route evolves its own skill library and memory from the same initial harness; the routes use the same coding model, observations, verifier, and budgets and differ only in what else the agent may use. Code calls code skills only and receives no demonstration. Code+ICL additionally receives a demonstration of the task in context and cannot call the policy. Code+Policy can additionally call the current policy with a short local instruction, and receives no demonstration. In every route, the agent's program is executed from the failure state or from the original reset; the failure state is restored by replaying the policy's recorded actions. Each repair work item contains the parent harness revision, the task and its initial state, the original goal and verifier, the localized failure, the affected skill, retrieved repair cases, and a trial budget. \Cref{fig:libero-routes} shows the three routes on one LIBERO-PRO case, \emph{turn off the stove}.

\paragraph{In-context demonstrations.}
On LIBERO-PRO, the Code+ICL demonstration of a task is an episode of a similar task from the official LIBERO training demonstrations, chosen without regard to outcome or quality. Frames similar to the current views are selected from both cameras and given with the recorded robot state and action at each frame. The agent is told that the demonstration shows a different scene and grounds every target in the current views.

\section{Repair Examples}

A validated repair is written back as a revision of a general skill in $L_r$ or, for policy orchestration, of the routing rules $\rho_r$ (\cref{sec:supp-skills}).

\begin{figure*}[t]
\centering
\repaircase{\includegraphics[width=\linewidth]{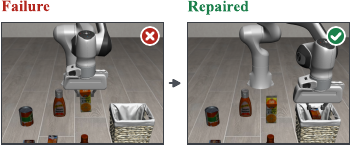}}{Task grounding and sequencing}{LIBERO-PRO, Code}{the program acts on the wrong object, or a step it needs is missing or out of order.}{skill arguments and the order of steps; add intermediate checks.}{target arguments and step order of any skill call.}{the first program grasps the orange juice instead of the butter; the revision grasps the butter.}
\repaircase{\includegraphics[width=\linewidth]{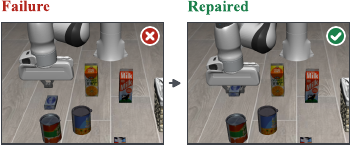}}{Grasp acquisition}{LIBERO-PRO, Code}{the object is not retained after closing.}{approach alignment, grasp depth, and orientation; add a VLM retention check.}{\texttt{bound\_grasp}.}{the first grasp closes above the cream cheese; a lower contact point grasps it.}
\repaircase{\includegraphics[width=\linewidth]{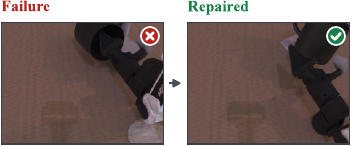}}{Collision-free motion}{RoboTwin}{the approach, carrying, or retreat path meets an obstacle or cannot be planned.}{add clearance waypoints; change approach, carrying, and retreat directions.}{approach and retreat phases of \texttt{bound\_grasp}, \texttt{bound\_place}, and \texttt{bound\_transfer}.}{the retreat after hanging the mug cannot be planned; a horizontal retreat away from the rack can.}
\repaircase{\includegraphics[width=\linewidth]{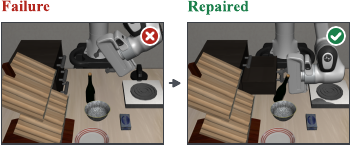}}{Contact and articulation}{LIBERO-PRO, Code}{contact with a handle, knob, or pushed object is lost or never established, as seen in the VLM check.}{the contact mode, for example an open-finger push, a hooked handle, or a stroke that keeps contact; stop when the VLM check shows progress.}{\texttt{bound\_drawer\_pull} and \texttt{bound\_surface\_stroke}.}{pulling with the fingers pointing down leaves the drawer closed; horizontal fingers behind the handle open it.}
\repaircase{\includegraphics[width=\linewidth]{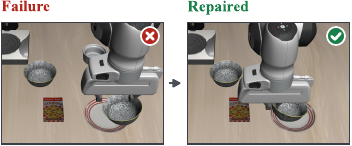}}{Placement and transitions}{LIBERO-PRO, Code}{the object is released off target, or the release and retreat leave the next step infeasible.}{placement offset, insertion direction, release, and withdrawal path.}{\texttt{bound\_place} and \texttt{bound\_transfer}.}{\emph{pick the black bowl next to the plate and place it on the plate}.}
\repaircase{\includegraphics[width=\linewidth]{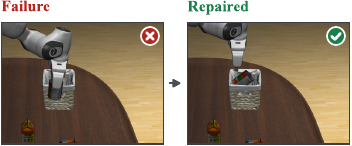}}{Temporal efficiency}{LIBERO-PRO, Code+ICL}{idle periods or slow phases use up the horizon.}{shorten idle transitions, bound speed changes, and fit the remaining phases into the remaining budget.}{per-phase step limits of the skills in the program.}{with 35 of 400 actions left, the revision finishes placing the sauce can in the basket.}
\repaircase{\includegraphics[width=\linewidth]{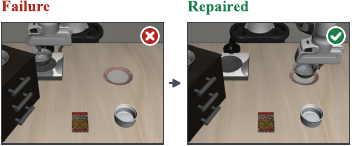}}{Policy orchestration}{LIBERO-PRO, Code+Policy}{in Code+Policy, a policy call fails or leaves the next step infeasible.}{the subgoal instruction, chunk length, retries, and which steps are routed to the policy.}{\texttt{vla\_act} calls and the routing rules $\rho_r$.}{the policy's grasp leaves the bowl on the stove; a code grasp replaces the call.}
\vspace{-4pt}
\caption{\textbf{Repair skills.} One recorded case per repair: the state after the failed program and the end of the repaired execution.}
\label{fig:supp-repair-skills}
\end{figure*}

\Cref{fig:supp-repair-skills} lists when each repair applies, what it changes, and what it edits; the revision is kept only if the complete task succeeds and earlier successes that use the edited skill still succeed.

\section{Policy Training Details}
\label{sec:supp-training}

\subsection{Training Setup and Data}

\Cref{tab:supp-cpd-recipe,tab:supp-recipe-robotwin,tab:supp-recipe-molmo,tab:supp-recipe-g1d} list the policy training setup on each benchmark and on the G1-D robot. Training data come from successful policy rollouts and from verified harness-generated trajectories of two kinds: full-task solutions from a matched task reset, and recovery continuations from the state where the policy failed, within the remaining horizon. In each group of executions that share a start state and source type, the smoothest success is kept (\cref{sec:supp-trajectory-quality}). Records store the source state, instruction, policy and harness versions, actions, remaining horizon, outcome, and cost.

\begin{table}[!htb]
\centering\small
\caption{\textbf{Policy training on LIBERO-PRO} ($\pi_{0.5}$).}
\label{tab:supp-cpd-recipe}
\begin{tabular}{@{}lr@{}}
\toprule
Setting & Value\\
\midrule
LoRA rank (VLM / action expert) & 512 / 512\\
Action chunk length & 10\\
Optimizer & AdamW\\
Updates per round / warmup & 30,000 / 500\\
Learning rate, cosine (peak / final) & $2.5\times10^{-5}$ / $2.5\times10^{-6}$\\
Gradient clipping & 1.0\\
$\beta$ (margin $m=1/(2\beta)$) & 0.5 (1.0)\\
Negative-loss allowance $\kappa$ & 0.3\\
$\lambda_{\rm pref}$ / $\lambda_{\rm prox}$ & 0.02 / 0.5\\
Failure-weight exponent $\gamma$ & 1\\
\bottomrule
\end{tabular}
\end{table}

\begin{table}[!htb]
\centering\small
\caption{\textbf{Policy training on RoboTwin C2R} ($\pi_{0.5}$).}
\label{tab:supp-recipe-robotwin}
\begin{tabular}{@{}lr@{}}
\toprule
Setting & Value\\
\midrule
Initialization & $\pi_{0.5}$ base\\
Fine-tuning & Full parameters\\
Training data (clean / autonomous) & 2.5K / 5K\\
Updates & 20,000\\
Batch size & 800\\
\bottomrule
\end{tabular}
\end{table}

\begin{table}[!htb]
\centering\small
\caption{\textbf{Policy training on MolmoSpaces} ($\pi_{0.5}$).}
\label{tab:supp-recipe-molmo}
\begin{tabular}{@{}lr@{}}
\toprule
Setting & Value\\
\midrule
Initialization & $\pi_{0.5}$\\
Updates & 5,000\\
Batch size & 1,376\\
Learning rate & $1\times10^{-5}$\\
\bottomrule
\end{tabular}
\end{table}

\begin{table}[!htb]
\centering\small
\caption{\textbf{Policy training on the Unitree G1-D} ($\pi_{0.5}$, one policy per task).}
\label{tab:supp-recipe-g1d}
\begin{tabular}{@{}lr@{}}
\toprule
Setting & Value\\
\midrule
Initialization & $\pi_{0.5}$ base\\
Fine-tuning & Full parameters\\
Updates per task & 10,000\\
Batch size & 16\\
Learning rate, cosine (peak / final) & $2.5\times10^{-5}$ / $2.5\times10^{-6}$\\
\bottomrule
\end{tabular}
\end{table}

\subsection{Motion Smoothness}
\label{sec:supp-trajectory-quality}

The normalized jerk of a trajectory $\tau$ is
\begin{equation}
 J(\tau)=\frac{D_\tau^5}{|\mathcal A_g|L_g^2}
 \sum_{a\in\mathcal A_g}\int_0^{D_\tau}
 \|\dddot p_a(t)\|_2^2\,dt,
\end{equation}
where $D_\tau$ is its duration, $p_a(t)$ the tool-center position of arm $a$, $\mathcal A_g$ the arms used by task $g$, and $L_g$ a fixed task scale; $J$ is unchanged by uniform time scaling.

\section{Evaluation Protocol}
\label{sec:supp-benchmarks}

Standalone success runs the policy alone, without the harness or any agent call; hybrid success runs the same policy with the harness, which routes each step to the policy or a code skill. \Cref{tab:supp-tasks} lists the evaluation tasks. \Cref{tab:supp-access} compares the information available to the agent in Harness VLA~\cite{zhang2026harness} and in RoboEvo$^2$. \Cref{tab:supp-real-config} lists the real-robot configuration.

\begin{table*}[t]
\centering\small
\caption{\textbf{Evaluation tasks.}}
\label{tab:supp-tasks}
\setlength{\tabcolsep}{5pt}
\renewcommand{\arraystretch}{1.1}
\begin{tabularx}{\textwidth}{@{}>{\raggedright\arraybackslash}p{0.12\textwidth}>{\raggedright\arraybackslash}p{0.17\textwidth}>{\raggedright\arraybackslash}X>{\raggedright\arraybackslash}p{0.21\textwidth}@{}}
\toprule
Benchmark & Robot & Tasks & Evaluation episodes\\
\midrule
LIBERO-PRO & Franka Panda & Spatial, Object, Goal, and Long suites, 10 tasks each, under position and task perturbations (8 settings) & Initial states 10--19 per task; 100 per setting, 800 in total\\
RoboTwin C2R & ALOHA AgileX & 50 bimanual tasks; evolution on 20 seeds per task not used in evaluation & 100 seeds per task; 5,000 in total\\
Cross-embodiment & ALOHA AgileX, dual ARX-X5, dual Franka-Panda & The 50 RoboTwin tasks with the frozen harness & 20 held-out layouts per task; 1,000 per robot\\
MolmoSpaces & Franka & Open, close, pick, and pick-and-place, 10 tasks each & 10 trials per task; 400 in total\\
Real world & Unitree G1-D & Stacking three bowls, relocating a block beside a container, opening a drawer and inserting a toy, retrieving a toy and closing the drawer & 10 trials per task and condition\\
\bottomrule
\end{tabularx}
\end{table*}

\begin{table}[!htb]
\centering\small
\caption{\textbf{Information available to the agent at evaluation.} Harness VLA entries follow its released code.}
\label{tab:supp-access}
\setlength{\tabcolsep}{4pt}
\begin{tabular}{@{}lcc@{}}
\toprule
Information & Harness VLA & RoboEvo$^2$\\
\midrule
RGB, proprioception, instruction & $\checkmark$ & $\checkmark$\\
Rendered depth, camera calibration & $\checkmark$ & $\checkmark$\\
Simulator object names (LIBERO) & $\checkmark$ & $\times$\\
Success flag after each primitive & $\checkmark$ & $\times$\\
Policy calls stopped by success flag & $\checkmark$ & $\times$\\
\bottomrule
\end{tabular}
\end{table}

\begin{table}[!htb]
\centering\small
\caption{\textbf{Real-robot configuration.}}
\label{tab:supp-real-config}
\setlength{\tabcolsep}{4pt}
\begin{tabularx}{\linewidth}{@{}l>{\raggedright\arraybackslash}X@{}}
\toprule
Component & Configuration\\
\midrule
Robot & Unitree G1-D, control model G1\_29\\
Arms & Two arms, 7 DoF each\\
Grippers & One Dex1 per arm\\
Depth camera & External Orbbec Gemini 336L; RGB $640\times480$ at 15\,FPS, depth $640\times400$ at 10\,FPS\\
Other cameras & Head stereo camera, $1280\times480$; one $640\times480$ camera per wrist\\
\bottomrule
\end{tabularx}
\end{table}

\section{Agent Prompts}
\label{sec:supp-prompt}

The agent uses two prompts: the execution prompt writes and revises programs from the current views, and the evolution prompt modifies the harness itself (skills, memory, and routing rules) from recorded failures.

\begin{promptblock}
You control a robot from the current camera images and the robot state.

## OBSERVATION

The instruction is the task; no object poses, simulator state, or rewards are available. Select targets as pixels in the current views and never invent world coordinates.

## SKILLS

Call only the listed skills, within their step and motion limits.

## ROUTE

<route>

## RESPONSE FORMAT

Before finishing, check every requested object and relation in the current views. Return one JSON object with a brief visual hypothesis, the skill calls, and the next step:

{"hypothesis": "<visual hypothesis>",
 "operations": [<skill calls>],
 "next": "reobserve" | "finish"}
\end{promptblock}

The ROUTE section of each route reads:

\begin{promptblock}
Code: Only code skills are available; no demonstration is attached and the policy cannot be called.

Code+ICL: The policy cannot be called. The demonstration shows a different scene; transfer its strategy, not its object locations.

Code+Policy: For local contact you may call the policy with one short instruction and a step budget, e.g.
{"op": "vla_act", "prompt": "grasp the visible white mug", "steps": 40}
Stage with code skills first and re-observe after the call.
\end{promptblock}

The user message is one JSON object followed by the current RGB and depth images of each camera view; the demonstration field is present only on Code+ICL:

\begin{promptblock}
{"instruction": "<instruction>",
 "cameras": {"<view>": <image size, intrinsics,
   camera-to-world>},
 "robot": <robot state>,
 "remaining_steps": <steps>,
 "remaining_revisions": <programs>,
 "skills": <skill catalog>,
 "general_lessons": <retrieved lessons>,
 "reusable_experiences":
   <retrieved experiences>,
 "previous_public_execution": <status of each
   call, executed actions, end-effector error,
   stop reason>,
 "training_demonstration":
   <demonstration frames>}
<view> current RGB: <image>
<view> current depth: <image>
\end{promptblock}

The evolution system prompt is:

\begin{promptblock}
You improve the harness from its own failures.

## DIAGNOSIS

Diagnose each failure from the recorded frames.

## CHANGES

Revise the affected skill or add a new one. Record what you learned as memory entries that cite the attempts supporting them. Edit the routing rules when a phase should move between the policy and code skills. Work on a private copy and test each change within the trial budget.

## ACCEPTANCE

A change is kept only if the complete task then succeeds and earlier successes that use the edited skills still succeed.

## RESPONSE FORMAT

Return a patch, a memory delta with its evidence, and the attempted trajectories.
\end{promptblock}

and its user message is:

\begin{promptblock}
Failure: <task, initial state, recorded frames, affected skill>
Harness: <skill code, memory, routing rules>
Similar repairs: <retrieved cases>
Budget: <trials>
\end{promptblock}

\section{Detailed Results}

\Cref{tab:supp-rounds} gives the values plotted in Fig.~3 of the main paper; \cref{tab:supp-libero-suite,tab:supp-robotwin-task} break down the final results on LIBERO-PRO and RoboTwin C2R.

\begin{table}[!htb]
\centering\small
\caption{\textbf{Success (\%) after each evolution round.} Round 0 is the initial policy and harness. RoboTwin C2R runs five rounds.}
\label{tab:supp-rounds}
\setlength{\tabcolsep}{4pt}
\begin{tabular}{@{}lrrrr@{}}
\toprule
& \multicolumn{2}{c}{LIBERO-PRO} & \multicolumn{2}{c}{RoboTwin C2R}\\
\cmidrule(lr){2-3}\cmidrule(l){4-5}
Round & Standalone & Hybrid & Standalone & Hybrid\\
\midrule
0 & 11.0 & 30.4 & 46.0 & 48.8\\
1 & 32.1 & 46.1 & 51.4 & 58.6\\
2 & 42.8 & 62.9 & 57.8 & 68.5\\
3 & 52.3 & 68.3 & 64.3 & 77.1\\
4 & 65.6 & 82.6 & 75.7 & 85.3\\
5 & 76.4 & 86.0 & 77.5 & 87.4\\
6 & 76.8 & 86.8 & -- & --\\
7 & 82.9 & 88.1 & -- & --\\
8 & 84.1 & 90.5 & -- & --\\
\bottomrule
\end{tabular}
\end{table}

\begin{table}[!htb]
\centering\small
\caption{\textbf{LIBERO-PRO success (\%) by suite} (RoboEvo$^2$, $\pi_{0.5}$ + PPO; 100 episodes each). Pos.\ and Task are the position and task perturbations.}
\label{tab:supp-libero-suite}
\setlength{\tabcolsep}{3.2pt}
\begin{tabular}{@{}lrrrrrr@{}}
\toprule
& \multicolumn{3}{c}{Standalone} & \multicolumn{3}{c}{Hybrid}\\
\cmidrule(lr){2-4}\cmidrule(l){5-7}
Suite & Pos. & Task & Avg. & Pos. & Task & Avg.\\
\midrule
Spatial & 98.0 & 97.0 & 97.5 & 99.0 & 100.0 & 99.5\\
Object & 100.0 & 98.0 & 99.0 & 100.0 & 98.0 & 99.0\\
Goal & 98.0 & 90.0 & 94.0 & 99.0 & 95.0 & 97.0\\
Long & 80.0 & 97.0 & 88.5 & 85.0 & 98.0 & 91.5\\
\midrule
All & 94.0 & 95.5 & 94.8 & 95.8 & 97.8 & 96.8\\
\bottomrule
\end{tabular}
\end{table}

\begin{table*}[t]
\centering\footnotesize
\caption{\textbf{RoboTwin C2R success (\%) by task} after five rounds, 100 seeds per task.}
\label{tab:supp-robotwin-task}
\setlength{\tabcolsep}{5pt}
\renewcommand{\arraystretch}{0.92}
\begin{tabular}{@{}lrrclrr@{}}
\toprule
Task & Standalone & Hybrid && Task & Standalone & Hybrid\\
\cmidrule(r){1-3}\cmidrule(l){5-7}
\texttt{adjust\_bottle} & 99 & 99 && \texttt{place\_can\_basket} & 61 & 68\\
\texttt{beat\_block\_hammer} & 73 & 89 && \texttt{place\_cans\_plasticbox} & 35 & 81\\
\texttt{blocks\_ranking\_rgb} & 76 & 85 && \texttt{place\_container\_plate} & 96 & 100\\
\texttt{blocks\_ranking\_size} & 61 & 72 && \texttt{place\_dual\_shoes} & 73 & 83\\
\texttt{click\_alarmclock} & 100 & 100 && \texttt{place\_empty\_cup} & 100 & 100\\
\texttt{click\_bell} & 100 & 100 && \texttt{place\_fan} & 71 & 91\\
\texttt{dump\_bin\_bigbin} & 92 & 95 && \texttt{place\_mouse\_pad} & 61 & 78\\
\texttt{grab\_roller} & 100 & 100 && \texttt{place\_object\_basket} & 64 & 73\\
\texttt{handover\_block} & 65 & 68 && \texttt{place\_object\_scale} & 78 & 96\\
\texttt{handover\_mic} & 97 & 99 && \texttt{place\_object\_stand} & 82 & 94\\
\texttt{hanging\_mug} & 31 & 34 && \texttt{place\_phone\_stand} & 75 & 88\\
\texttt{lift\_pot} & 98 & 100 && \texttt{place\_shoe} & 98 & 98\\
\texttt{move\_can\_pot} & 57 & 81 && \texttt{press\_stapler} & 98 & 100\\
\texttt{move\_pillbottle\_pad} & 57 & 79 && \texttt{put\_bottles\_dustbin} & 85 & 86\\
\texttt{move\_playingcard\_away} & 94 & 98 && \texttt{put\_object\_cabinet} & 53 & 63\\
\texttt{move\_stapler\_pad} & 50 & 59 && \texttt{rotate\_qrcode} & 97 & 97\\
\texttt{open\_laptop} & 96 & 99 && \texttt{scan\_object} & 71 & 89\\
\texttt{open\_microwave} & 64 & 71 && \texttt{shake\_bottle} & 99 & 100\\
\texttt{pick\_diverse\_bottles} & 70 & 91 && \texttt{shake\_bottle\_horizontally} & 97 & 100\\
\texttt{pick\_dual\_bottles} & 85 & 94 && \texttt{stack\_blocks\_three} & 61 & 66\\
\texttt{place\_a2b\_left} & 71 & 93 && \texttt{stack\_blocks\_two} & 95 & 98\\
\texttt{place\_a2b\_right} & 71 & 91 && \texttt{stack\_bowls\_three} & 76 & 77\\
\texttt{place\_bread\_basket} & 68 & 91 && \texttt{stack\_bowls\_two} & 93 & 96\\
\texttt{place\_bread\_skillet} & 68 & 77 && \texttt{stamp\_seal} & 75 & 95\\
\texttt{place\_burger\_fries} & 86 & 97 && \texttt{turn\_switch} & 52 & 91\\
\bottomrule
\end{tabular}
\end{table*}

\section{Ablations}
\label{sec:supp-agent-ablation}

We replace GPT-6 Astra with Qwen3.8-27B~\cite{qwen2026qwen38_27b}. We run the same evolution rounds as the main LIBERO-PRO run from the same initial policy and harness, and only the agent changes. \Cref{tab:supp-agent-ablation} reports standalone and hybrid success after the final round.

\par\medskip\noindent
\begin{minipage}{\linewidth}
\centering\small
\captionsetup{type=table}
\caption{\textbf{Agent-backbone ablation on LIBERO-PRO.} Success (\%) after the final round.}
\label{tab:supp-agent-ablation}
\setlength{\tabcolsep}{3pt}
\begin{tabularx}{\linewidth}{@{}Xrr@{}}
\toprule
Agent & Standalone & Hybrid\\
\midrule
GPT-6 Astra & 84.1 & 90.5\\
Qwen3.8-27B & 76.0 & 82.5\\
\bottomrule
\end{tabularx}
\end{minipage}
\par\medskip

We also compare CPD with and without EWC ($\lambda_{\rm EWC}=0$), keeping the training data, initialization, adapters, and optimizer budget fixed. Besides LIBERO-PRO, we evaluate the updated policy on the original LIBERO benchmark~\cite{liu2023libero} to measure forgetting (\cref{tab:supp-ewc-ablation}).

\par\medskip\noindent
\begin{minipage}{\linewidth}
\centering\small
\captionsetup{type=table}
\caption{\textbf{EWC ablation.} Standalone success (\%) on LIBERO and LIBERO-PRO.}
\label{tab:supp-ewc-ablation}
\setlength{\tabcolsep}{3pt}
\begin{tabularx}{\linewidth}{@{}Xrr@{}}
\toprule
Policy update & LIBERO & LIBERO-PRO\\
\midrule
CPD + EWC & 92.5 & 84.1\\
CPD, $\lambda_{\rm EWC}=0$ & 72.0 & 84.3\\
\bottomrule
\end{tabularx}
\end{minipage}
\par\medskip

To measure run-to-run variation, we repeat the full LIBERO-PRO run three times with the same configuration, each for eight rounds (\cref{tab:supp-repeat-runs}).

\par\medskip\noindent
\begin{minipage}{\linewidth}
\centering\small
\captionsetup{type=table}
\caption{\textbf{Repeated runs on LIBERO-PRO.} Standalone policy success (\%) after eight rounds.}
\label{tab:supp-repeat-runs}
\setlength{\tabcolsep}{3pt}
\begin{tabularx}{\linewidth}{@{}>{\centering\arraybackslash}X>{\centering\arraybackslash}X>{\centering\arraybackslash}X>{\centering\arraybackslash}X@{}}
\toprule
Run 1 & Run 2 & Run 3 & Mean $\pm$ std\\
\midrule
84.1 & 82.1 & 84.9 & 83.7 $\pm$ 1.4\\
\bottomrule
\end{tabularx}
\end{minipage}
\par\medskip

\section{Limitations}

RoboEvo$^2$ composes and revises skills on top of a fixed set of atomic skills and perception tools, which do not yet cover capabilities such as force-controlled contact or in-hand regrasping. Future work can enlarge this initial library with such primitives, which the agent can then compose and revise like the existing ones. A task-level success check selects the executions that train the policy; the VLM check that repaired skills already use could replace it, at the cost of extra model calls.